\documentclass[dvipsnames,table,xcdraw, 10pt,twocolumn,letterpaper]{article}

\usepackage{iccv}              

\usepackage{multirow}
\usepackage{graphicx}
\usepackage{caption}
\usepackage[pagebackref,breaklinks,colorlinks,allcolors=iccvblue]{hyperref}
\usepackage{subcaption}
\usepackage[symbol]{footmisc}

\definecolor{iccvblue}{rgb}{0.21,0.49,0.74}

\def\paperID{2256} 
\def\confName{ICCV}
\def\confYear{2025}

\title{Generative Active Learning for Long-tail Trajectory Prediction via \\ Controllable Diffusion Model}

\author{
Daehee Park$^{1}$\textsuperscript{\dag}, Monu Surana$^{2}$, Pranav Desai$^{2}$, Ashish Mehta$^{2}$, Reuben MV John$^{2}$, and Kuk-Jin Yoon$^{3}$ \\
$^1$Intelligent Systems and Learning Lab., DGIST, Korea \\
$^2$Qualcomm Research, USA 
$^3$Visual Intelligence Lab., KAIST, Korea 
}

\begin{document}
\maketitle
\begin{abstract}
While data-driven trajectory prediction has enhanced the reliability of autonomous driving systems, it still struggles with rarely observed long-tail scenarios.
Prior works addressed this by modifying model architectures, such as using hypernetworks.
In contrast, we propose refining the training process to unlock each model’s potential without altering its structure.
We introduce \textbf{G}enerative \textbf{A}ctive \textbf{L}earning for \textbf{Traj}ectory prediction (\textbf{GALTraj}), the first method to successfully deploy generative active learning into trajectory prediction.
It actively identifies rare tail samples where the model fails and augments these samples with a controllable diffusion model during training.
In our framework, generating scenarios that are diverse, realistic, and preserve tail-case characteristics is paramount. 
Accordingly, we design a tail-aware generation method that applies tailored diffusion guidance to generate trajectories that both capture rare behaviors and respect traffic rules.
Unlike prior simulation methods focused solely on scenario diversity, GALTraj is the first to show how simulator-driven augmentation benefits long-tail learning in trajectory prediction. 
Experiments on multiple trajectory datasets (WOMD, Argoverse2) with popular backbones (QCNet, MTR) confirm that our method significantly boosts performance on tail samples and also enhances accuracy on head samples.
\end{abstract}    
\footnotetext[2]{ Work done during an internship at Qualcomm Research.}
\section{Introduction}
\label{sec:intro}

Predicting the future motion of dynamic traffic agents is crucial in autonomous systems. 
Recent data-driven methods~\cite{alahi2016social, lee2017desire, gupta2018social, rhinehart2018r2p2, ivanovic2019trajectron, bae2023set} have achieved remarkable success in complex, interactive scenarios \cite{xu2024adapting, pourkeshavarz2024cadet, bae2024can, cheng2023forecast, seff2023motionlm}, and state-of-the-art predictors now attain high accuracy on large-scale real-world datasets such as nuScenes~\cite{caesar_nuscenes_2020} and Argoverse~\cite{chang_argoverse_2019}. 
Despite these advances, they remain vulnerable to the \emph{long-tail problem}, failing on rarely observed \emph{tail} samples~\cite{Wang_2023_CVPR, mercurius2024amend, lian2025cdkformer}. 
This arises because data-driven models bias their representations toward frequently seen (head) samples, leaving underrepresented (tail) samples insufficiently modeled.
Although existing prediction benchmarks gauge performance primarily on major (head) data, the safety-critical nature of autonomous systems makes accurate prediction of rare tail cases indispensable~\cite{makansi2021exposing}.

\begin{figure}[t]
\centering
\centerline{\includegraphics[width=0.95\columnwidth]{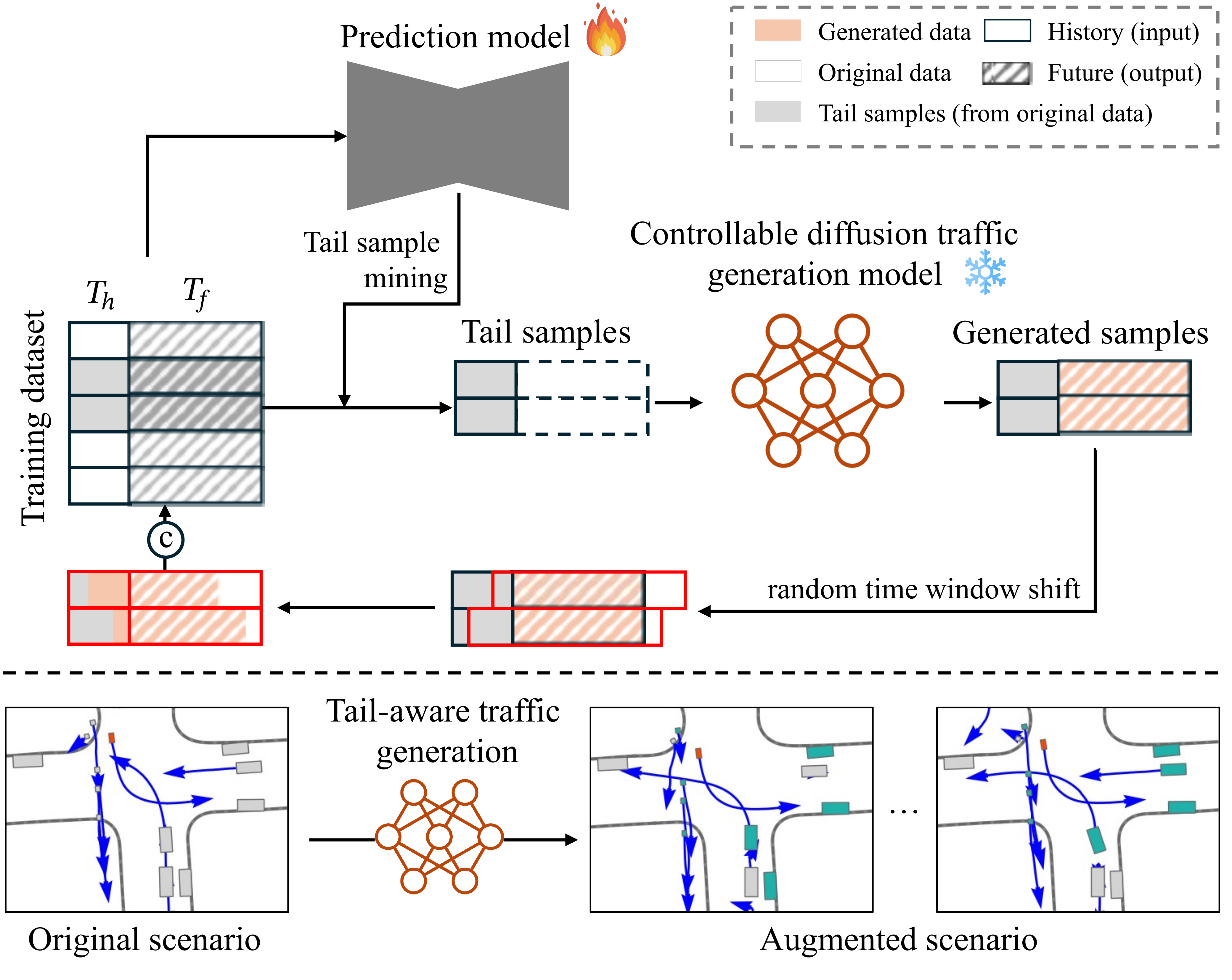}}
\vspace{-2pt}
\caption{
\textbf{Overview of our method.}
 Each sample in the dataset corresponds to a traffic scenario involving multiple interacting agents. In each training epoch, we identify tail samples with high prediction errors and augment them using our tail-aware generation method. This yields realistic yet diverse scenarios that preserve tail characteristics, thereby mitigating data imbalance. Notably, this is the first work to harness a generative traffic simulator to address the long-tail problem in trajectory prediction.}
\label{fig:enter-label}
\vspace{-5pt}
\end{figure}

The long-tail problem has been extensively studied in computer vision and machine learning~\cite{Ouyang_2016_CVPR,NEURIPS2020_1e14bfe2,Samuel_2021_ICCV}, where it is typically framed as a class imbalance: head classes have many samples, tail classes few~\cite{li2022targeted}.
However, it also arises in regression tasks like trajectory prediction~\cite{yang2021delving}, since rare driving behaviors (e.g., U-turns, sudden overtakes) are underrepresented. 
Recent works have tackled this problem by modifying network architectures (e.g., adding hypernetworks or expert modules)~\cite{makansi2021exposing, mercurius2024amend}; however, these methods increase model complexity and introduce additional hyperparameters (e.g., the number of clusters), which can degrade performance on head samples~\cite{Wang_2023_CVPR}. 

To this end, we propose \emph{changing the training procedure} instead of modifying the backbone network. 
As illustrated in Fig.~\ref{fig:enter-label}, we propose a \textbf{generative active learning} framework for trajectory prediction: at each iteration, it dynamically identifies tail samples, augments them, and updates the training dataset in each iteration. 
This is achieved by leveraging a \emph{controllable diffusion traffic simulator} to generate new future trajectories. 
While traffic simulation~\cite{jiang2024scenediffuser, huang2024versatile} has been used to diversify scenarios, this is the first work demonstrating that simulator-driven data generation can improve long-tail performance in trajectory prediction.

However, naively simulating random traffic scenes does not solve the long-tail imbalance.
We therefore design a \emph{tail-aware} generation method that accounts for the interacting nature of traffic scenes.
It is designed to generate scenarios that preserve the characteristics of tail samples while ensuring scene-level diversity and traffic rule constraints.
Specifically, we categorize traffic agents into tail, head, and relevant groups, then assign distinct guidance within the diffusion model. 
The proposed augmentation and learning strategy lead to enhanced prediction performance on both tail samples and the entire dataset.
We validate our method on multiple popular benchmarks (WOMD, Argoverse 2) and with different backbone models (QCNet, MTR), consistently observing larger gains than baseline methods.

\noindent We summarize our contributions as follows:
\begin{itemize}
    \item  
    We introduce a generative active learning for the trajectory prediction task using a controllable traffic generator, marking the first approach to show traffic simulation can successfuly benefit long-tail learning for trajectory prediction.
    
    \item 
    We propose a tail-aware generation method that assigns distinct guidance to each agent category, enabling the generation of realistic and diverse tail scenarios while preserving crucial tail behaviors.

    \item 
    Our approach is validated on multiple datasets and backbones, demonstrating not only remarkable improvement on tail samples but also on the entire dataset.
\end{itemize}
\section{Related works}
\label{sec:related}

\subsection{Trajectory prediction}
Trajectory prediction is essential for autonomous systems operating in multi-agent environments.
By forecasting the future states of surrounding agents from historical data, these systems enable safe and efficient path planning~\cite{chen2025ppad, huang2023gameformer, phong2024truly, hu2023planning, lee2025non}.
Recent data-driven approaches have significantly improved long-term prediction accuracy~\cite{jeong2024multi}, surpassing traditional rule-based approaches.
Accurate prediction requires capturing inter-agent interactions, agent-environment dynamics, and the multi-modal nature of future trajectories~\cite{bae2023eigentrajectory, mu2024most, bae2022learning}.
To address these challenges, recent works have explored advanced architectures such as transformers, diffusion models, and graph neural networks~\cite{tang2024hpnet, wang2024optimizing, jiang2023motiondiffuser, rowe2023fjmp, Bae_2024_CVPR}.
Some focus on modeling complex agent interactions~\cite{wen2024density, kim2024higher, park2023leveraging, Bae_2024_CVPR_language, xu2025sports}, while others enhance scene understanding by incorporating environmental context~\cite{gu2024producing, lan2023sept, zhang2024real, jeong2025multi}.
Beyond architecture design, researchers are also addressing the limitations of datasets~\cite{park2024t4p, zhou2024smartrefine, park2024improving}, particularly the long-tail problem caused by data imbalance.
Efforts to mitigate this issue have recently gained attention, emphasizing the need to improve prediction reliability in rare but critical scenarios.

\subsection{Long-tail learning}
The long-tail problem arises when a small number of dominant (head) classes overshadow rare (tail) classes, leading to biased models that perform poorly on tail data~\cite{cao2019learning, menon2021longtail, liu2019openlongtail, cao2019ldam}. 
Existing solutions fall into three categories: class re-balancing, information augmentation, and module improvements~\cite{10105457, zhou2020bbn, alshammari2022long, xu2021towards}.
Class re-balancing methods adjust the distribution of training samples~\cite{jamal2020rethinking, menon2021logitadjust, zhong2021improving, hou2023subclass}, while information augmentation techniques, such as transfer learning and data augmentation, provide additional data or features~\cite{li2021metasaug}. 
Module improvement strategies refine network architectures to enhance robustness~\cite{li2023meid, wang2021rsg, wang2021contrastive, cui2021parametric, li2022nested}.
Trajectory prediction datasets also suffer from long-tail issues, as rare driving scenarios like U-turns or risky overtakes are underrepresented. 
Recent studies have attempted to address this by modifying model architectures, for example, using hypernetworks or mixtures of experts~\cite{Wang_2023_CVPR, mercurius2024amend}. 
However, such approaches increase model complexity and introduce additional hyperparameters (e.g., the number of clusters), which may degrade performance on head samples.

\subsection{Information augmentation in long-tail learning}
Information augmentation techniques, including transfer learning and data augmentation, introduce additional information to improve learning~\cite{yang2020rethinking, he2021distilling}. 
Transfer learning enables knowledge transfer from a source to a target domain, enabling models to be pre-trained on long-tail samples and fine-tuned on balanced subsets or vice versa~\cite{cui2018large, yang2020rethinking}.
Data augmentation enhances tail class diversity at both the feature and data levels. 
Feature-level augmentation methods, such as FTL~\cite{yin2019feature} and LEAP~\cite{liu2020deep}, aim to reduce the intra-class variance within tail classes.
Data-level augmentation approaches~\cite{zhou2023imbsam}, like M2m~\cite{kim2020m2m}, generate tail samples by transforming head class instances. 
More recent techniques~\cite{zang2021fasa, li2021metasaug} synthesize diverse yet semantically consistent tail data, improving performance without sacrificing head class accuracy.
Several studies have further enhanced augmentation strategies by incorporating active learning~\cite{kong2019active, huang2024active, zhang2024expanding}. 
However, these methods primarily focus on image classification and are not directly applicable to trajectory prediction, a multi-agent regression task. 
To bridge this gap, we propose a generative active learning framework specifically designed for trajectory prediction.
\section{Method}
\label{sec:method}
\subsection{Problem definition}
Trajectory prediction aims to estimate agents' future positions $\textup{\textbf{y}} = \left\{ x^n_t, y^n_t \right\}_{\Delta t : T_f}^{1:N}$ from their past observations $ \textup{\textbf{x}} = \left\{ x^n_t, y^n_t \right\}_{-T_h : 0}^{1:N}$.
Here, $n$, $t$, and $N$ represent the agent index, time index, and the number of agents within a scene, respectively. 
$T_f$, $T_h$, and $\Delta t$ denote the future horizon, observation horizon, and time interval.
Trajectory datasets consist of traffic scenarios, represented as $\mathcal{D} = \left\{ S_j \right\}_{ \left | \mathcal{D} \right |}$,
where each $S_j$ is the $j$th scenario $ \left\{ \textup{\textbf{x}}, \textup{\textbf{y}} \right\}$.
Our objective is to train a predictor $\psi$ on the training dataset $\mathcal{D}_{tr}$ so that it performs effectively on both tail and head samples of the validation dataset $\mathcal{D}_{vl}^{all}$ .
\subsection{Overall method}

The proposed method follows an active learning framework~\cite{kong2019active, huang2024active}.
We begin by training the prediction model on the original dataset following the backbone models’ standard procedure, stopping at two-thirds of the total training epochs.
This initial training is essential, as identifying meaningful tail samples is difficult when training from scratch.
In the next step, we identify tail samples where the trained model fails, allowing us to detect data patterns that the model finds challenging (Sec.~\ref{sec:tail_mining}).
We then augment tail samples through \textit{tail-aware} generation (Sec.~\ref{sec:tail_aware_generation}).
Using the augmented data, we establish an iterative training loop to enhance model performance (Sec.~\ref{sec:training_loop}).
The details of each step are outlined below.

\subsection{Tail sample mining}
\label{sec:tail_mining}
Accurate identification of tail samples is essential to our method.
Previous methods detect tail samples using clustering~\cite{mercurius2024amend, Wang_2023_CVPR} or by measuring errors with a Kalman filter~\cite{makansi2021exposing}.  
However, these approaches are suboptimal, as they do not capture the actual failure cases of the target prediction model.  
Clustering assumes that small groups correspond to tail samples, but this does not always imply high prediction error. 
Similarly, errors from Kalman filters do not necessarily reflect the actual errors of the target model.

In contrast, our method defines tail samples dynamically, where the prediction model at the current epoch fails to make accurate predictions.  
For each agent $n$ at epoch $e$, we compute the prediction error $\delta^{n, (e)}$.  
An agent is classified as a tail agent if its error exceeds a threshold $\tau$; the corresponding scene is then marked as a tail sample. 
The set of tail samples $\mathcal{D}_{tr}^{tail, (e)}$ within the training dataset $\mathcal{D}_{tr}$ is defined as:
\begin{align}
    & \mathcal{D}_{tr}^{tail, (e)} = \left\{ S_j \in \mathcal{D}_{tr} \mid \max_{n \in S_j} \delta^{n, (e)} > \tau \right\}, \\
    & \text{where} \quad \delta^{n, (e)} = \text{error}(\psi^{(e)}(\textbf{x}^n), \textbf{y}^n).
\end{align}
We use minADE$_6$ as the prediction error metric throughout our method.
Once tail samples are identified, their scenario IDs and agent IDs are stored in memory.
Note that the per-agent prediction error is already computed during the original model’s loss calculation, eliminating the need for an additional inference pass.  
The only additional step is to threshold the prediction error and record the IDs of tail agents and scenes.

\subsection{Tail-aware generation method}
\label{sec:tail_aware_generation}
We use a pretrained generative diffusion model $\Theta$ to augment identified tail samples by generating future trajectories $\boldsymbol{\hat{\mathrm{y}}}$ from past observations \(\textbf{\textup{x}}\) taken from tail scenarios  $S_j \in \mathcal{D}_{tr}^{tail, (e)}$.
While various methods exist for traffic scenario generation~\cite{pronovost2023scenario, tan2023language}, generating arbitrary scenarios without careful design is unlikely to be effective for long-tail learning.
To address this, we design a \textit{tail-aware} generation method that ensures the generated samples meaningfully contribute to long-tail learning.
There are two key considerations for generating data that truly benefits long-tail learning.
First, the generated scenes must be both diverse and representative of tail sample characteristics.
Second, the generated scenes must be realistic (i.e., socially compliant and adhering to traffic rules), as unrealistic training samples can lead to learning irrelevant features, degrading overall performance.

\subsubsection{Generation with real guidance}

\begin{figure}[t]
    \centering
    \centerline{\includegraphics[width=0.95\columnwidth]{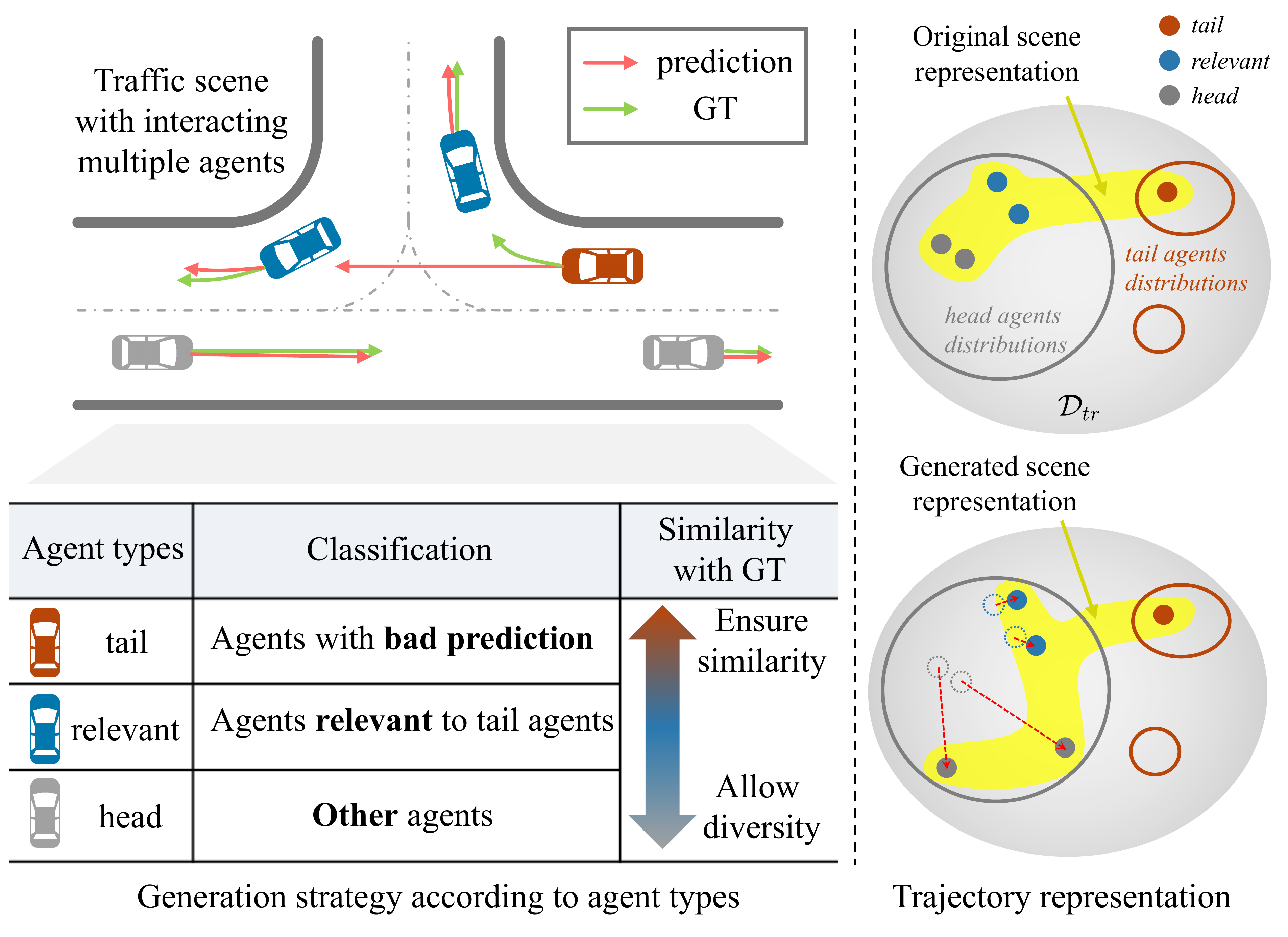}}
    \caption{Agent type categorization and corresponding generation strategy. We allow varying diversity based on agent types. This strategy maintains the structural characteristics of tail samples while diversifying scene composition, ensuring that generated tail samples effectively mitigate data imbalance. In trajectory representation, head, relevant, and tail agents move progressively less in the generated scene compared to the original scene. However, the overall scene-level representation undergoes significant changes, which have a greater impact on learning.}
    \label{fig:agent_category}
\end{figure}

\paragraph{Generation strategy.}
The first key consideration is preserving semantic similarity with tail samples while allowing diversity in generation~\cite{li2024semantic}.
Unlike image classification tasks, where each data sample is a single, independent entity that can be generated in a class-conditioned manner, traffic scenarios exhibit different characteristics.
They consist of multiple interacting agents, each influencing the overall scene dynamics.
Applying conditioned generation to traffic scenarios without accounting for agent interactions can lead to unrealistic and unstructured samples, failing to capture the complexities of tail samples.
Thus, a more structured approach is required.

Since we define tail samples as scenes where the prediction model fails, agents within the scene can be categorized based on specific criteria, as illustrated in the left part of Fig.~\ref{fig:agent_category}.
First, based on the prediction error, agents are classified into either \textit{tail} or \textit{head} agents.
We define \textit{tail} agents as those for which the model fails to make accurate predictions, whereas \textit{head} agents are those it successfully predicts.
In scenario generation, preserving the motion characteristics of \textit{tail} agents is crucial for maintaining the essence of tail samples.
Conversely, introducing diverse motion patterns for \textit{head} agents enhances scenario variety, making the generated scenes more effective in the training process.
In other words, the motions of \textit{tail} agents should closely resemble their ground-truth future trajectories, while \textit{head} agents should exhibit greater variation by deviating from their original trajectories to introduce scene-level diversity.

However, excessive motion diversity in all \textit{head} agents can lead to implausible scenarios.
For instance, a generated motion for a \textit{head} agent may result in a collision with a \textit{tail} agent, creating unrealistic interactions.
To mitigate this, we introduce an additional classification for certain \textit{head} agents that significantly interact with \textit{tail} agents; we refer to them as \textit{relevant} agents.
\textit{Relevant} agents are identified using an \textbf{agent-agent interaction module} within the diffusion decoder, which determines interaction strength based on attention scores. 
Agents whose attention score exceeds $\frac{1}{|\mathcal{N}_j|}$, where $\mathcal{N}_j$ denotes the set of neighboring agents, are classified as relevant.
With this three-category classification, we generate scenarios by assigning different levels of diversity to each agent type.

As shown on the right side of Fig.~\ref{fig:agent_category}, when multiple agents interact within an original scene, applying different levels of diversity to each agent results in new traffic scene compositions.
Notably, diversity among \textit{head} agents plays an important role because the trajectory encoder considers all agents in a scene when computing their interaction representations.
Increasing scene-level diversity enhances the representation of tail samples, leading to a broader distribution of learned features.

\begin{figure}[t]
    \centering
    \centerline{\includegraphics[width=0.95\columnwidth]{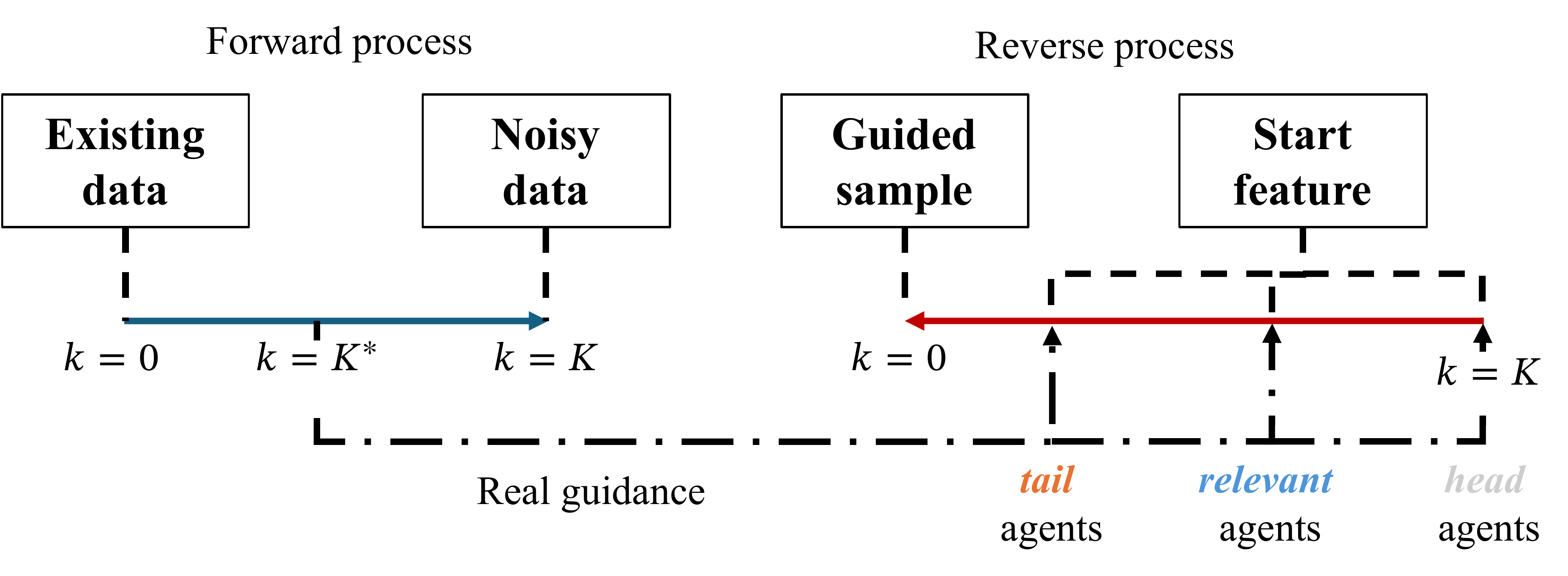}}
    \vspace{-3pt}
    \caption{
    Visualization of real guidance to assign different levels of similarity and diversity in generation. 
    Under real guidance, model samples from noised ground truth rather than pure noise, resulting in samples that resemble the ground truth. 
    We assign a different starting point $K^*$ to each agent type.}
    \label{fig:method_real_guidance}
\end{figure}

\paragraph{Control of diversity.}
To control the diversity of generated trajectories, we apply real guidance within the generation process~\cite{he2023is}, as shown in Fig.~\ref{fig:method_real_guidance}.
The standard diffusion generation process begins with random noise and iteratively denoises it through diffusion steps for $k = K \rightarrow 0$.
Although the generation is guided by the log-likelihood learned from the entire dataset, the resulting samples tend to follow the dominant modes of the data distribution.
As a result, rare or long-tail samples, which occupy low-probability regions of the distribution, are unlikely to be generated.
Real guidance addresses this limitation by initializing the reverse process not from random noise at $k=K$, but from the noised ground truth at an intermediate step, obtained via the forward process.
The reverse process then starts from the time step $K^*$. 
By adjusting $K^*$, we can control the similarity between the generated samples and the ground-truth distribution.

We set $K^*$ to progressively higher values for \textit{tail}, \textit{relevant}, and \textit{head} agents.
These values are empirically determined for each agent type as follows:
\begin{align}
    & p(\boldsymbol{\hat{\mathrm{y}}}) \approx p_{\theta}(\boldsymbol{\hat{\mathrm{y}}}_0 \mid \boldsymbol{\mathrm{y}}_{K^*}, \boldsymbol{\mathrm{x}}), \quad K^* = \lambda_{\textit{type}} K,
\end{align}
where $\lambda_{\textit{type}}$ is a scaling factor that varies based on the agent type with empirically chosen values of $\lambda_{\textit{tail}} = 0.25$, $\lambda_{\textit{rel}} = 0.6$, and $\lambda_{\textit{head}} = 1$.
As a larger $K^*$ corresponds to a noisier starting point, it allows for greater diversity in generation.

\subsubsection{Generation with gradient guidance}

In traffic scenarios, agents generally adhere to traffic rules.
While \textit{tail} and \textit{relevant} agents are guided by real constraints, \textit{head} agents are not directly constrained, which may result in the violation of traffic rules.
To mitigate this issue, we apply gradient-based guidance during inference for \textit{head} agents, encouraging compliance with traffic regulations, following~\cite{zhong2023guided}.
This method perturbs the predicted mean at each denoising step using the gradient of a predefined objective function, $\mathcal{C}$, directly modifying the mean at the current step.
The process is formulated as follows:
\begin{equation}
p_\theta\left(\textbf{\textup{y}}_{k-1} \mid \textbf{\textup{y}}_k, \boldsymbol{\mathrm{x}}\right) \approx \mathcal{N}\left(\boldsymbol{\textbf{y}}_{k-1} ; \boldsymbol{\mu}+\boldsymbol{\Sigma}^k \nabla_{\boldsymbol{\mu}} \mathcal{C}(\boldsymbol{\mu}), \boldsymbol{\Sigma}^k\right).
\end{equation}
We enforce two traffic rules: the \textit{no-off-road}, which ensures that generated trajectories stay within road boundaries, and the \textit{repeller}, which prevents collisions between generated trajectories. 
For detailed mathematical formulations, please refer to the supplementary material.

\subsection{Training loop with overfitting mitigation}
\label{sec:training_loop}
A generated scenario is represented as:
\begin{equation}
    S_j^{'} = \left\{ \boldsymbol{\mathrm{x}}, \boldsymbol{\hat{\mathrm{y}}} \right\} = \left\{ \textup{\textbf{p}}^n_t \right\}_{-T_h : T_f}^{1:N}.
\end{equation}
Since each generated scenario spans only the future prediction horizon, output features vary while input features remain fixed.
To address this, we introduce a simple yet effective technique, \textbf{random time window shift}, for each generated scenario:
\begin{equation}
    S_j^{''} = \left\{ \textup{\textbf{p}}^n_t \right\}_{-T_h+\delta t : T_f+\delta t}^{1:N}.
\end{equation}
This ensures that a portion of the generated future trajectories is used as historical context, thereby diversifying input features and mitigating overfitting.
Details on how $\delta t$ is selected are provided in the supplementary material.
For time steps beyond the generated horizon ($T_f : T_f + \delta t$), positions are zero-padded and masked during training.

The augmented inputs and outputs are then concatenated into the training dataset to update it:
\begin{equation}
    \mathcal{D}_{tr}^{(e+1)} = \mathcal{D}_{tr}^{(e)} \cup \mathcal{D}_{tr}^{gen, (e)}, \quad \mathcal{D}_{tr}^{gen, (e)} = \left\{ S_j^{''} \right\}.
\end{equation}
To ensure newly generated \emph{tail} scenarios are more frequently sampled during training, we decay the sampling weights of the previous epoch’s dataset by a factor $\alpha$, while assigning a weight of 1 to the new data.
Sampling weights are clipped at a predefined minimum to retain sufficient coverage of head data and prevent performance degradation caused by overfitting to generated scenes.
Post-training is then performed using the updated training dataset.
Finally, the iterative training loop consists of tail sample mining, generation, dataset updates, and post-training.
Note that tail sample mining is conducted only on the original training dataset, excluding generated scenes. 
Since the generated samples are derived from the original dataset, including them may lead to redundant detection of tail samples.

\section{Experiments}
\label{sec:experiment}
We evaluate our method using multiple backbone models: QCNet~\cite{zhou2023query} and MTR~\cite{shi2022motion}.
We use the official implementation of QCNet, and MTR is obtained from the UniTraj~\cite{feng2024unitraj} repository.
We use the WOMD~\cite{ettinger_large_2021} and Argoverse2~\cite{wilson2021argoverse} datasets.
All agents in the scene are predicted and evaluated, as our setting considers the entire scene.
For diffusion-based traffic generation, we adopt LCSim~\cite{zhang2024lcsim}.
In our training procedure, for fair comparison, we use an identical number of training data samples per epoch across all methods, implemented through fixed-size random sampling.
More details on the datasets, backbone models, and diffusion generation model are provided in the supplementary material.

\subsection{Baselines}

\noindent We compare our method with various learning paradigms:

\noindent\textbf{Vanilla}: The standard training procedure without any modifications. This corresponds to the original prediction model and serves as a direct baseline.

\noindent\textbf{Resampling}~\cite{shi2023re}: Unlike classification tasks, tail samples are not explicitly defined in regression tasks; we identify them using a pretrained prediction model and assign higher sampling weights. 
Unlike our approach, this method does not involve data generation; instead, it directly increases the sampling frequency of identified tail samples at the end of each epoch following standard re-sampling techniques in long-tail learning.

\noindent\textbf{cRT}~\cite{Kang2020Decoupling}: A decoupled approach where feature encoders are first trained with uniform sampling, then fixed while the decoder is re-trained on a balanced dataset. Following the resampling baseline, no generative augmentation is applied.

\noindent\textbf{Contrastive}~\cite{makansi2021exposing}: We compare with an open-source long-tail learning method for trajectory prediction that uses a contrastive loss to pull representations of challenging samples closer in the feature space.
This helps the model learn more discriminative representations for tail samples.

\noindent\textbf{Naive}: A straightforward adaptation of generative active learning to trajectory prediction. Tail samples are identified and augmented, then directly incorporated into the training dataset without the proposed tail-aware generation method.

\subsection{Evaluation}
We evaluate both long-tail and overall prediction performance.
We use minADE$_6$ per agent as the standard metric for evaluating long-tail performance.

\noindent\textbf{Top k\%} error is a long-tail metric that represents the prediction error for the k\% most challenging samples, as identified by the pre-trained prediction model~\cite{Wang_2023_CVPR}. 
It indicates how well the prediction model adapts to tail samples.

\noindent \textbf{Value-at-risk} (\textbf{VaR$_\alpha$}) is another long-tail metric that quantifies the magnitude of the error distribution of the current model~\cite{mercurius2024amend}.
Defined as the $\alpha^{th}$ quantile of the error distribution, it measures performance on the worst-performing samples, reflecting how favorable the error distribution of the model is.
Unlike Top k\%, which measures performance improvement on pre-identified tail samples, VaR evaluates the error distribution of the current model itself.

\noindent \textbf{False prediction ratio} (\textbf{FPR$_{th}$}) is also a long-tail metric that measures the percentage of false predictions on the whole dataset.
A false prediction is defined as an agent with a prediction error greater than the threshold.
This also reflects how favorable the error distribution of the model is.

\noindent \textbf{minADE$_6$ (all)} and \textbf{minFDE$_6$ (all)} are overall metrics.
They are widely used metrics in trajectory prediction that measure the minimum distance between predicted multi-modal trajectories and the ground truth future trajectory over the entire prediction horizon or at the final time step.
Both metrics are averaged across the entire dataset, and for simplicity, the notation (all) will be omitted from here on.
\section{Results}
\label{sec:result}
\subsection{Model capacity}
\begin{table}[t]
\caption{This experiment tests our assumption that refining the training procedure can unlock the model’s potential.
We check whether the model has sufficient complexity to represent both head and tail samples without architecture modification.
}
\label{tab:result_model_cap}
\centering
\resizebox{\columnwidth}{!}{
\begin{tabular}{l|cccc}
\toprule
Method     &  \begin{tabular}[c]{@{}c@{}}Top 1\% \end{tabular} & \begin{tabular}[c]{@{}c@{}}VaR$_{999}$\end{tabular} & \begin{tabular}[c]{@{}c@{}}FRR$_5$\end{tabular} & minADE$_6$         \\ \hline
Pretrained & 7.38          & 10.04         & 0.73                     & 0.374        \\
GALTraj   & \textbf{2.29} & \textbf{3.02} & \textbf{0.12}            & \textbf{0.272} \\ \bottomrule
\end{tabular}
}
\end{table}
Our primary assumption is that recent prediction models possess the capacity to capture both head and tail scenarios, yet suboptimal training limits their long-tail performance. 
To validate this, we trained QCNet using our method on the WOMD training split and evaluated it on the same split.
As shown in Tab.~\ref{tab:result_model_cap}, our approach yields significant improvements on all long-tail metrics, confirming that existing architectures can accommodate tail data without additional modules. 
This finding underscores that even state-of-the-art predictors are often constrained by their training procedures (see Sec.~\ref{sec:main_results} for cross-split generalizability).

\subsection{Generation results}
\begin{figure}[t]
    \centering
    \centerline{\includegraphics[width=0.99\columnwidth]{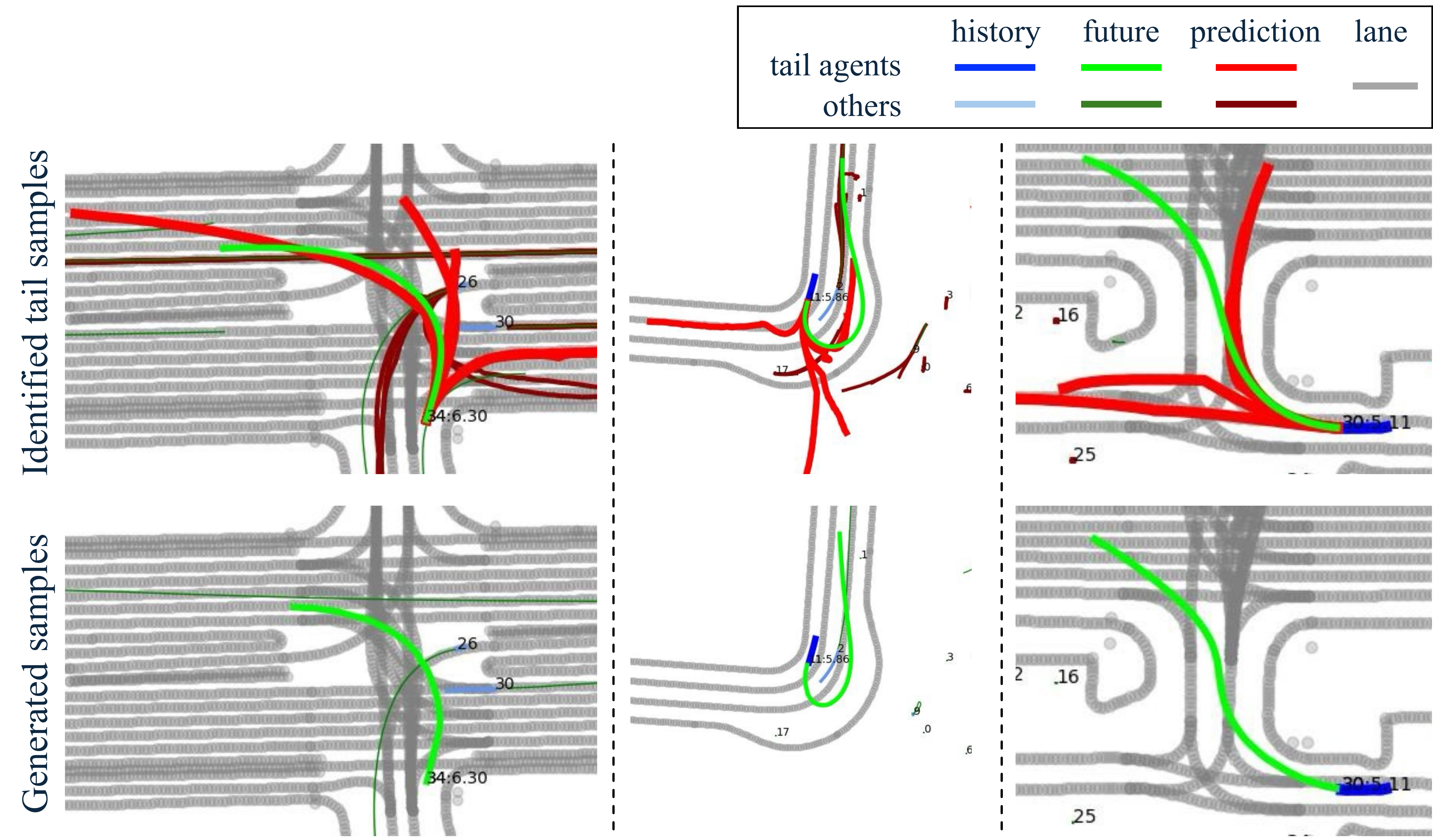}}
    \caption{
    Visualization of tail sample mining (top) and tail-aware generation (bottom). 
    In the top row, the prediction model fails to accurately forecast tail agents’ trajectories. 
    In the bottom row, the generated trajectories closely resemble the ground-truth future for tail agents while maintaining distinct variations.
    }
    \label{fig:viz_hard_gen}
\end{figure}
 
The first row of Fig.~\ref{fig:viz_hard_gen} shows the identified tail samples, with tail agents highlighted by thick, bright lines.
The model struggles most when the road structure is complex, with multiple possible directions, and on rare maneuvers like U-turns.
The second row presents traffic scenarios generated from these tail samples using our tail-aware generation method.
Thanks to tailored diffusion guidance, the generated trajectories closely match the ground truth while introducing distinct scenario variations.

\begin{table}[t]
\caption{Main experimental results. 
The backbone prediction model (QCNet) is trained using various training methods and compared across them.
Both long-tail and overall metrics are measured.
Lower values indicate better performance for all metrics.}
\label{tab:main_result}
\resizebox{\columnwidth}{!}{
\centering
\begin{tabular}{c|l|ccc|c}
\toprule
                          &                                  & \multicolumn{3}{c|}{Long-tail metrics}                                                       & \begin{tabular}[c]{@{}c@{}}Overall \\ metric\end{tabular} \\ \cline{3-6} 
\multirow{-2}{*}{} & \multirow{-2}{*}{Method}         & Top 1\%                      & VaR$_{999}$                     & FPR$_5$                           & minFDE$_6$                                                   \\ \hline
                          & Vanilla                          & 4.81                         & 8.42                         & 0.42                         & 0.654                                                     \\
                          & resampling                       & 4.30                         & 8.01                         & 0.38                         & 0.668                                                     \\
                          & cRT                              & 4.45                         & 8.42                         & 0.43                         & 0.645                                                     \\
                          & contrastive                      & 4.12                         & 6.71                         & 0.31                         & 0.613                                                     \\
                          & Naive                            & 4.56                         & 7.91                         & 0.38                         & 0.612                                                     \\
\multirow{-6}{*}{\rotatebox[origin=c]{90}{WOMD}}    & \cellcolor[HTML]{EFEFEF}GALTraj & \cellcolor[HTML]{EFEFEF}\textbf{3.43} & \cellcolor[HTML]{EFEFEF}\textbf{6.05} & \cellcolor[HTML]{EFEFEF}\textbf{0.22} & \cellcolor[HTML]{EFEFEF}\textbf{0.558}                             \\ \hline
                          & Vanilla                          & 4.47                         & 7.22                         & 0.35                         & 0.545                                                     \\
                          & resampling                       & 4.04                         & 6.86                         & 0.28                         & 0.571                                                     \\
                          & cRT                              & 4.12                         & 7.05                         & 0.29                         & 0.547                                                     \\
                          & contrastive                      & 3.92                         & 5.97                         & 0.23                         & 0.544                                                     \\
                          & Naive                            & 4.40                         & 6.95                         & 0.32                         & 0.530                                                     \\
\multirow{-6}{*}{\rotatebox[origin=c]{90}{AV2}}     & \cellcolor[HTML]{EFEFEF}GALTraj & \cellcolor[HTML]{EFEFEF}\textbf{3.76} & \cellcolor[HTML]{EFEFEF}\textbf{5.66} & \cellcolor[HTML]{EFEFEF}\textbf{0.19} & \cellcolor[HTML]{EFEFEF}\textbf{0.524}                             \\ \bottomrule
\end{tabular}
}
\end{table}

\subsection{Main results}
\label{sec:main_results}
\paragraph{Quantitative results.}
Table~\ref{tab:main_result} shows that GALTraj improves QCNet’s performance across both WOMD and Argoverse2. 
Our method delivers substantial gains on all long-tail metrics.
Most notably, FPR$_5$ is reduced by half, indicating a substantial reduction in extreme prediction errors.
\begin{figure*}[t]
    \centering
    \centerline{\includegraphics[width=0.99\linewidth]{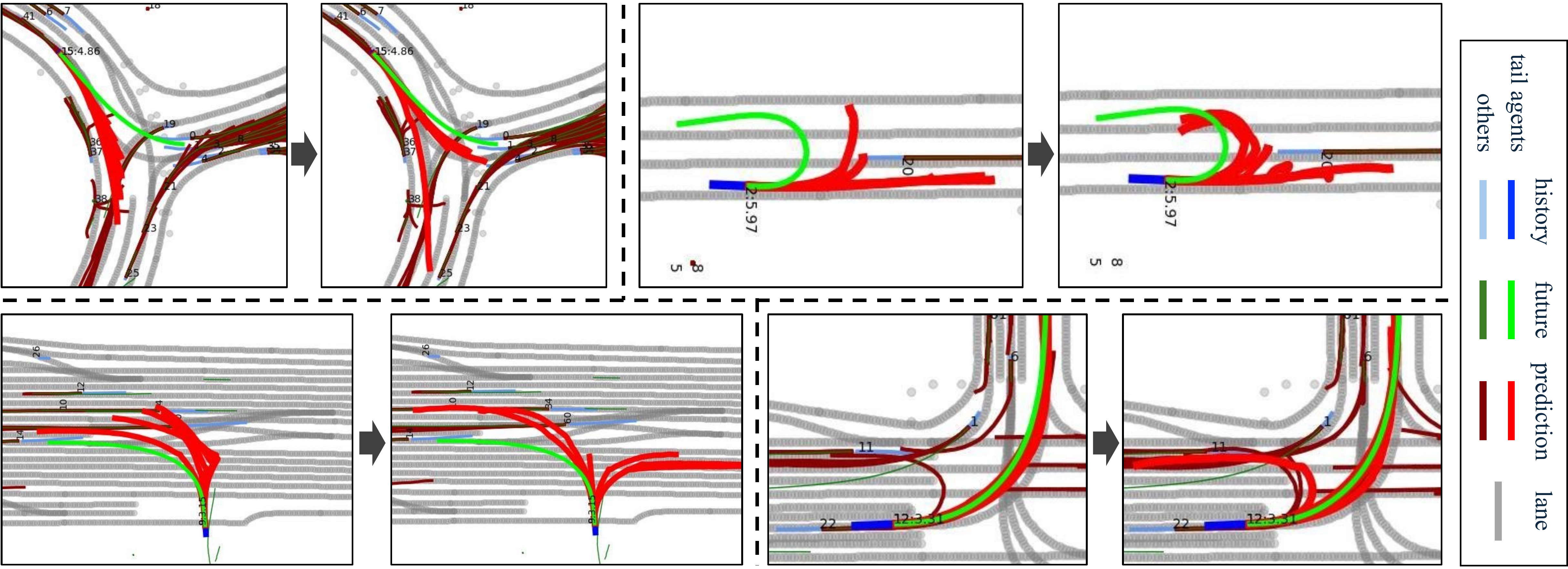}}
    \caption{
    Visualization of main experimental results.
    The left side of the image pair shows the model trained with the \textit{vanilla} method, while the right side shows the model trained with the proposed method.
    It shows that the proposed method predicts unique motions even in challenging scenarios and learns a more diverse future representation.
    }
    \label{fig:viz_main_compare}
\end{figure*}

Moreover, GALTraj improves overall metrics, while baseline methods sometimes worsen minFDE$_6$, as also observed in FEND~\cite{Wang_2023_CVPR}.
This degradation is caused by overfitting to irrelevant context due to simple concatenation of tail samples, as in the \textbf{resampling} method.
By contrast, the proposed augmentation method produces diverse, realistic trajectories that enrich feature learning and drive robust overall improvements.

The \textbf{Naive} method yields only modest gains, highlighting the critical role of our tail-aware generation strategy.
In summary, GALTraj not only adapts effectively to rare tail scenarios but also maintains strong generalization across the entire dataset. 
For a deeper dive into error distributions, please see the supplementary material.

\begin{table}[t]
\caption{Experiments with MTR backbone on WOMD. 
This experiment tests the generalizability of our method to different backbone models.}
\label{tab:mtr_exp}
\centering

\begin{tabular}{l|ccc|c}
\toprule
                         & \multicolumn{3}{c|}{Long-tail metrics}         & \begin{tabular}[c]{@{}c@{}}Overall \\ metric\end{tabular} \\ \cline{2-5} 
\multirow{-2}{*}{Method} & Top 1\%       & VaR$_{999}$         & FPR$_5$          & minFDE$_6$                                                   \\ \hline
Vanilla                  & 7.71          & 15.95          & 0.99          & 0.806                                                     \\
resampling               & 7.02          & 14.58          & 0.87          & 0.823                                                     \\
cRT                      & 7.22          & 15.14          & 0.93          & 0.798                                                     \\
contrastive              & 6.75          & 12.81          & 0.74          & 0.780                                                     \\
Naive                    & 7.70          & 15.38          & 0.96          & 0.794                                                     \\
\rowcolor[HTML]{EFEFEF} 
GALTraj                  & \textbf{5.87} & \textbf{12.03} & \textbf{0.65} & \textbf{0.773}                                            \\ \bottomrule
\end{tabular}
\vspace{-10pt}
\end{table}

\paragraph{Additional experiments.}
We further evaluate our approach using another popular backbone, MTR.
Table~\ref{tab:mtr_exp} shows that our method consistently outperforms baseline methods.
This finding confirms that our method generalizes well to multiple prediction backbones.
Further experiments using additional datasets (nuScenes~\cite{caesar_nuscenes_2020}), other metrics (3\%, 5\%, FRR$_{10}$) are included in the supplementary material.

\paragraph{Qualitative results.}
Figure~\ref{fig:viz_main_compare} provides qualitative results of the proposed method. 
In each image pair, the left side shows predictions from the model trained with the vanilla method, while the right side shows predictions from the model trained with our method. 
In the top row, we observe that while the vanilla method fails in complex environments or uncommon maneuvers, the proposed method successfully captures these challenging scenarios.
This demonstrates the proposed method’s ability to learn tail samples more effectively, producing more accurate predictions in rare but critical situations.
The bottom row reveals that our method captures a broader range of modalities, effectively representing diverse potential trajectories.
This capability aligns better with the multi-modal nature of trajectory prediction tasks, allowing the model to anticipate risks in uncertain environments and respond to varied possible scenarios.
More qualitative results, including classification results for different agent types, are provided in the supplementary material.

\subsection{Ablation studies}

\begin{table}[t]
\caption{Ablation experiments on four key components of the proposed method: real/gradient guidance, sampling weight decay, and random time-window shift. Results are from WOMD dataset.}
\label{tab:ablation_1}
\resizebox{\columnwidth}{!}{
\centering
\renewcommand{\arraystretch}{1.2}
\begin{tabular}{cl|ccccc}
\toprule
\multicolumn{1}{l}{}      &                   & \multicolumn{5}{c}{exp no.}                                                                       \\
\multicolumn{1}{l}{}      &                   & 1                    & 2                    & 3                    & 4                    & 5     \\ \hline
\multirow{4}{*}{\rotatebox[origin=c]{90}{components}} & Real guidance     & \multicolumn{1}{l}{} & \checkmark                    & \multicolumn{1}{l}{} & \checkmark                    & \checkmark     \\
                          & Gradient guidance & \multicolumn{1}{l}{} & \checkmark                    & \checkmark                    & \multicolumn{1}{l}{} & \checkmark     \\
                          & Sampling weight   & \multicolumn{1}{l}{} & \multicolumn{1}{l}{} & \checkmark                    & \checkmark                    & \checkmark     \\
                          & Random time shift & \multicolumn{1}{l}{} & \multicolumn{1}{l}{} & \checkmark                    & \checkmark                    & \checkmark     \\ \hline
\multirow{3}{*}{\rotatebox[origin=c]{90}{metrics}}  & FRR$_5$              & 0.38                 & 0.28                 & 0.34                 & 0.26                 & \textbf{0.22}  \\
                          & VaR$_{999}$           & 7.91                 & 6.49                 & 7.56                 & 6.52                 & \textbf{6.05}  \\
                          & minFDE$_6$           & 0.612                & 0.604                & 0.586                & 0.601                & \textbf{0.558} \\ \bottomrule
\end{tabular}
}
\end{table}

We conduct ablation studies on four main components of the proposed method: real guidance, gradient guidance, sampling weight decay, and random time-window shift.

In experiment 1, samples are generated using the generative model without any guidance, and these samples are naively concatenated into the training set. 
The generative model introduces diversity into the data samples, resulting in a slight performance improvement over the vanilla method.
However, the performance gain is limited.

In experiment 2, we observe that applying real and gradient guidance leads to significant performance improvements in long-tail metrics, such as FPR and VaR. 
Additionally, comparing experiment 2 with experiment 5, we find that the addition of the proposed sampling weight decay and random time-window shift not only further enhances long-tail metrics but also improves learning stability for head samples, resulting in overall performance improvements across all metrics.

\begin{figure}[t]
    \centering
    \centerline{\includegraphics[width=\columnwidth]{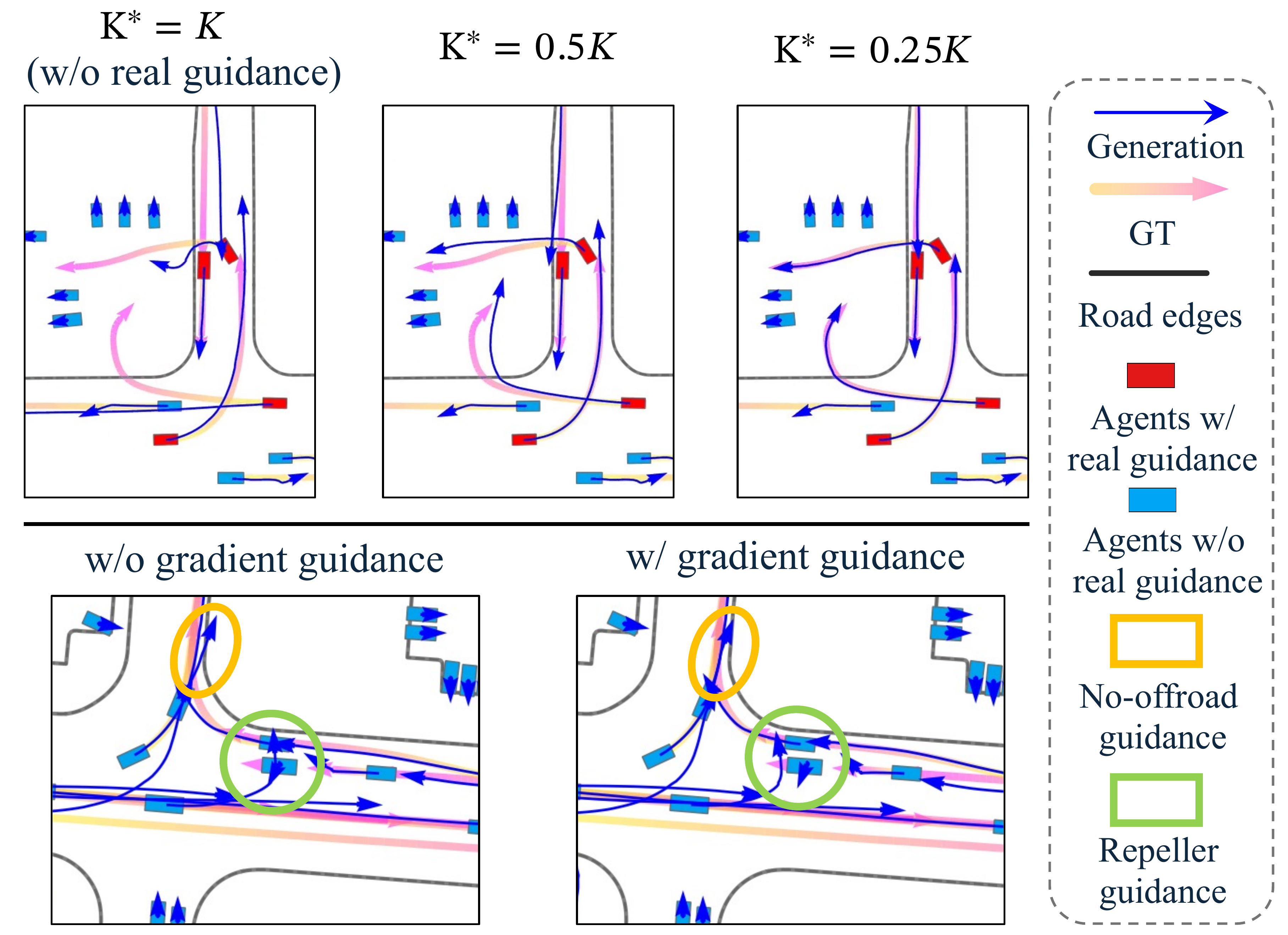}}
    \caption{
    Generation results according to gradient guidance.
    This guidance helps generate realistic scenarios by ensuring that the generated motion follows predefined traffic rules.
    }
    \label{fig:guidance_viz}
\end{figure}

Experiments 3 and 4 highlight the importance of the guidance in the tail-aware generation method. 
Comparing experiments 3 and 5, removing real guidance leads to a considerable decline in long-tail performance, underscoring the importance of real guidance in preserving characteristics of tail sample data.
The effect of real guidance is visualized in Fig.~\ref{fig:guidance_viz} (top row), showing that agents with real guidance preserve challenging behaviors, preventing oversimplified generation.
This finding indicates that real guidance is essential for the effective functioning of the proposed generative active learning framework, ensuring that the generated tail samples accurately capture the challenging characteristics needed to improve long-tail performance.

Comparing experiments 4 and 5, removing gradient guidance slightly degrades long-tail metrics but significantly worsens overall metrics.
This suggests that gradient guidance helps generate realistic scenarios and prevents performance degradation for head samples. 
Figure~\ref{fig:guidance_viz} (bottom row) illustrates this effect: without gradient guidance, generated trajectories frequently violate road constraints or overlap unrealistically with other agents.
By contrast, applying gradient guidance ensures that generated motions adhere to predefined traffic rules, such as off-road avoidance and collision prevention.
This leads to improved performance across head samples.

\subsection{Computational cost}
\label{sec:comput_cost}
The proposed method is applied only during offline training and does not modify the backbone network, so it does not impact inference time, which is crucial for real-time performance.
Nonetheless, we analyze the additional computation required during offline training. 
Because tail samples are identified using regression errors already computed during the prediction loss calculation, no additional forward passes are required.
A simple thresholding step combined with scene/agent ID hashing is sufficient.
The main computational overhead arises from generating novel samples based on identified tail samples. 
In our experiments, the maximum proportion of identified tail samples is less than 5\% of the training dataset.
As training converges, that share declines further, so the maximum additional training time per epoch is less than 36\% across all datasets and backbone models. 
This overhead also diminishes over time as fewer tail samples are identified with training convergence.
It could be further mitigated by adopting faster diffusion sampling methods in future work.
\section{Conclusion}
\label{sec:conclusion}
In this work, we address the long-tail problem in trajectory prediction by introducing a generative active learning framework.
Our method is the first to successfully leverage a generative traffic simulator to address the long-tail problem in trajectory prediction.
Instead of modifying the model architecture, we enhance training by identifying tail samples and subsequently generating targeted samples to directly mitigate data imbalance.
The proposed tail-aware generation method, based on a controllable diffusion model, significantly contributes to long-tail learning by augmenting diverse and realistic traffic scenarios while explicitly preserving the unique behaviors of tail samples.
Our experiments, conducted across multiple backbone models and datasets, demonstrate that our approach not only improves performance on challenging tail scenarios but also enhances overall prediction accuracy.
Future work may explore extending our generative active learning framework to related challenges in autonomous driving, such as motion planning.


\section*{Acknowledgment}
This work was supported by the Institute of Information \& Communications Technology Planning \& Evaluation(IITP) grant funded by the Korea government(MSIT) (No. RS-2025-02219277, AI Star Fellowship Support Project(DGIST)), and by the Institute of Information \& Communications Technology Planning \& Evaluation (IITP) grant funded by the Korea government(MSIT) (No. RS-2024-00457882, AI Research Hub Project).

{
    \small
    \bibliographystyle{ieeenat_fullname}
    \bibliography{main}
}

\end{document}